%% file: main.tex
\documentclass[10pt,twocolumn,letterpaper]{article}
\ifdefined\pdfsuppressptexinfo\pdfsuppressptexinfo=-1\fi
\usepackage[pagenumbers]{cvpr}
\def\confName{CVPR}
\def\confYear{2027}

\usepackage[T1]{fontenc}
\usepackage{graphicx}
\usepackage{amsmath,amssymb}
\usepackage{booktabs}
\usepackage{array}
\usepackage{multirow}
\usepackage{xcolor}
\usepackage{enumitem}
\usepackage{latexml}
\definecolor{cvprblue}{rgb}{0.21,0.49,0.74}
\usepackage[pagebackref,breaklinks,colorlinks,allcolors=cvprblue]{hyperref}
\usepackage[capitalize]{cleveref}
\usepackage{dblfloatfix}
\usepackage{cuted}
\ifdefined\captionof\else\usepackage{capt-of}\fi
\hypersetup{pdftitle={EEG-World: Context-Conditioned Dynamics of the Watching Brain},
  pdfauthor={Junxiang Zhang, Jiayu Zuo, Wenhao Ye, Junfeng Xia, Xiang Chen, Mo Wang, Quanying Liu}}

\newcommand{\method}{EEG-World}
\newcommand{\dataname}{VAEEG}

\title{\method: Context-Conditioned Dynamics of the Watching Brain}
\iflatexml
\author{Junxiang Zhang\thanks{Equal contribution.}\\
Southern University of Science and Technology, Shenzhen, China\\
Shenzhen Loop Area Institute, Shenzhen, China
\and
Jiayu Zuo\thanks{Equal contribution.}\\
Southern University of Science and Technology, Shenzhen, China\\
Omni Intelligence, Shenzhen, China
\and
Wenhao Ye\\
Shenzhen University, Shenzhen, China
\and
Junfeng Xia\\
Southern University of Science and Technology, Shenzhen, China
\and
Xiang Chen\\
Southern University of Science and Technology, Shenzhen, China\\
Omni Intelligence, Shenzhen, China
\and
Mo Wang\thanks{Co-corresponding author: 12250099@mail.sustech.edu.cn}\\
Southern University of Science and Technology, Shenzhen, China\\
The University of Warwick, Coventry, UK
\and
Quanying Liu\thanks{Co-corresponding author: liuqy@sustech.edu.cn}\\
Southern University of Science and Technology, Shenzhen, China\\
Omni Intelligence, Shenzhen, China\\
Shenzhen Loop Area Institute, Shenzhen, China}
\else
\author{%
Junxiang Zhang\textsuperscript{1,3,*}\quad
Jiayu Zuo\textsuperscript{1,2,*}\quad
Wenhao Ye\textsuperscript{4}\quad
Junfeng Xia\textsuperscript{1}\\
Xiang Chen\textsuperscript{1,2}\quad
Mo Wang\textsuperscript{1,5,\dag}\quad
Quanying Liu\textsuperscript{1,2,3,\dag}\\[6pt]
{\normalsize\textsuperscript{1}Southern University of Science and Technology, Shenzhen, China\quad
\textsuperscript{2}Omni Intelligence, Shenzhen, China}\\
{\normalsize\textsuperscript{3}Shenzhen Loop Area Institute, Shenzhen, China\quad
\textsuperscript{4}Shenzhen University, Shenzhen, China}\\
{\normalsize\textsuperscript{5}The University of Warwick, Coventry, UK}%
}
\fi
\date{}

\begin{document}
\maketitle
\iflatexml\else
\begin{NoHyper}
\renewcommand{\thefootnote}{*}\footnotetext{Equal contribution.}
\renewcommand{\thefootnote}{\dag}\footnotetext{\raggedright Co-corresponding authors: \mbox{12250099@mail.sustech.edu.cn}, \mbox{liuqy@sustech.edu.cn}}
\end{NoHyper}
\fi

\input{sections/abstract}
\input{sections/intro}

\input{sections/related}
\input{sections/data}
\input{sections/method}
\input{sections/benchmark}
\input{sections/experiments}
\input{sections/discussion}
\input{sections/llm}

{\small
\bibliographystyle{ieeenat_fullname}
\bibliography{refs}
}

\clearpage
\appendix
\setcounter{table}{0}
\setcounter{figure}{0}
\renewcommand{\thetable}{S\arabic{table}}
\renewcommand{\theHtable}{S\arabic{table}}
\renewcommand{\thefigure}{S\arabic{figure}}
\renewcommand{\theHfigure}{S\arabic{figure}}
\input{sections/appendix}

\end{document}

%% file: sections/abstract.tex
\begin{abstract}
Brain encoding models predict neural responses from audiovisual input, but typically overlook how ongoing brain activity shapes an individual's response. We introduce \textbf{EEG-World}, a recurrent world model that forecasts 10 seconds of EEG band-power dynamics from 6 seconds of preceding EEG and the subsequent audiovisual stimulus. Pretrained on continuous EEG, EEG-World learns shared neural dynamics and adapts to unseen individuals through lightweight subject-specific readouts.
Using 2,584 hours of EEG from 78 participants, we show that nonlinear recurrent forecasting surpasses strong history-aware linear baselines at scale. EEG-World achieves a forecasting correlation of 0.261 versus 0.233 for the strongest history-and-video linear model, with a 3.3-second forecasting-horizon advantage. Its learned dynamics transfer to unseen participants, outperforming the linear baseline in 23 of 24 individuals after calibration. While ongoing EEG history accounts for most predictive performance, controlled stimulus analyses reveal an additional video-dependent component. This component also enables evaluation of visual representations, with V-JEPA 2 improving clip-locked prediction by 35\% over combined CLIP and DINOv2 features. Our results demonstrate the value of transferable neural dynamics for individual EEG forecasting and highlight the importance of separating state-driven from stimulus-driven predictability.

\end{abstract}

%% file: sections/intro.tex
\section{Introduction}
\label{sec:intro}

\begin{figure}[t]
  \centering
  \includegraphics[width=\columnwidth]{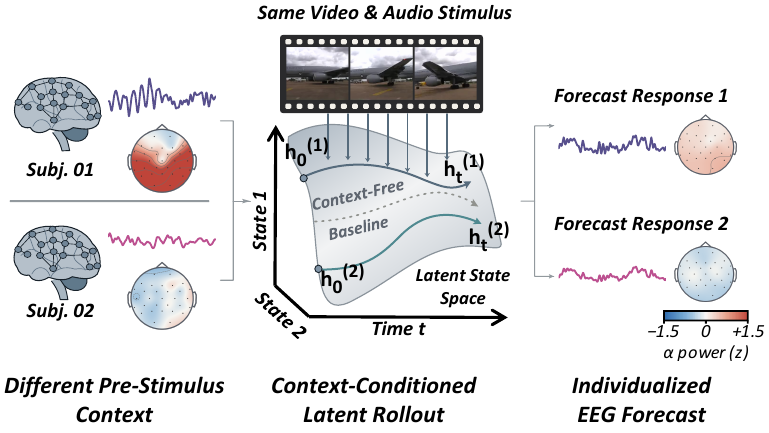}
  \caption{\textbf{Same clip, different viewers.} Different subjects start the same clip in different brain states
  (left). \method{} rolls each viewer's state forward under the clip and forecasts that viewer's own response (right);
  a context-free encoding model predicts the same trajectory for everyone (dashed). Traces are drawn from the forecast
  band power.}
  \label{fig:teaser}
\end{figure}

Brain recordings during naturalistic video viewing have become an important testbed for vision models: representations that predict neural activity are taken as evidence of alignment with human visual processing~\cite{yamins2014performance,schrimpf2018brainscore,huh2024platonic}. EEG makes such measurements more accessible and offers a temporally resolved view of brain activity. Predicting the neural response evoked by a stimulus, however, is not the same as forecasting how a particular viewer's brain activity will unfold while watching it. Even for the same video, different viewers can follow different neural trajectories because they start from different brain states (\cref{fig:teaser}). Forecasting these trajectories therefore requires modeling not only what is on screen but also the brain's ongoing dynamics.

Conventional stimulus-to-brain \emph{encoding models} predict neural responses from visual or audiovisual features~\cite{naselaris2011encoding,kay2008identifying,nishimoto2011reconstructing,huth2016natural}. Although effective at characterizing stimulus-related responses, they typically do not condition on an individual's preceding neural activity. This omission matters particularly for EEG, where ongoing rhythms and temporal persistence make recent activity highly predictive of the next few seconds. As a result, a model can forecast EEG accurately while learning little about the stimulus itself. This raises two questions: can nonlinear models capture neural dynamics beyond what strong history-aware linear models already explain, and can the resulting state-driven predictability be separated from the contribution of the video?

We approach this problem through the lens of \emph{world models}, which predict how a system evolves from its current state under external inputs~\cite{ha2018world,hafner2019planet,hafner2023dreamerv3}. In our setting, the brain is the dynamical system, EEG provides partial observations of its state, and the video acts as an exogenous input. We introduce \method{}, which infers a latent brain state from preceding EEG and rolls it forward under audiovisual conditioning to forecast an individual's neural activity. A recurrent dynamics core is pretrained on continuous EEG, and lightweight subject-specific readouts adapt the shared dynamics to new individuals. Because neural history alone explains much of the future activity, we compare against strong linear forecasters that receive the same EEG history and video, rather than against stimulus-only encoding models.

Learning and evaluating such dynamics requires data that existing EEG resources rarely combine: extensive recordings to learn temporal regularities, the same videos shown to many viewers to separate stimulus-driven from state-driven activity, and enough participants to hold some out for testing transfer. To meet these requirements, we collect VAEEG, a dataset of 2,584 hours of EEG from 78 participants watching naturalistic audiovisual clips. On this dataset, nonlinear recurrent forecasting overtakes strong linear baselines once hundreds of hours are pooled and keeps improving as linear models saturate. Its dynamics transfer to unseen participants, where 33 minutes of calibration match a linear model fitted on ten times more data. Although ongoing EEG dynamics dominate the predictive performance, controlled stimulus analyses reveal an additional video-dependent component whose in-silico tuning matches recorded EEG, enabling more informative comparisons between visual representations and human neural responses.

\paragraph{Contributions.}
\begin{itemize}[leftmargin=*,itemsep=1pt,topsep=2pt]
\item We introduce \method{}, a recurrent world model for individual EEG forecasting that combines pretrained neural dynamics, audiovisual conditioning, and lightweight subject-specific adaptation.
\item We demonstrate that nonlinear recurrent forecasting surpasses strong history-aware linear baselines at scale, with learned dynamics transferring to unseen participants, using 2,584 hours of naturalistic EEG from 78 participants.
\item We establish history-aware baselines and persistence-controlled evaluations that distinguish neural-state predictability from stimulus-driven responses, enabling more targeted comparisons of visual representations with the brain.
\end{itemize}

%% file: sections/related.tex
\section{Related work}
\label{sec:related}

\paragraph{Brain encoding models.}
Encoding models predict fMRI, MEG or EEG responses from stimulus features, usually with regularized linear
maps~\cite{naselaris2011encoding,kay2008identifying,nishimoto2011reconstructing,huth2016natural,allen2022nsd}, and how
well a vision model's features predict the brain is used to compare it with human vision~\cite{yamins2014performance,
schrimpf2018brainscore}. Recent challenges and models predict whole-brain fMRI responses to movies from multimodal
features~\cite{gifford2025algonauts,dascoli2025tribe}; for EEG and video, most work decodes the stimulus from the
response~\cite{liu2024eeg2video} or compares the response with the representations of vision
models~\cite{sartzetaki2026}. All of these treat the response as a function of the stimulus alone. They cannot say
what a particular brain will do next, because the ongoing state that shapes the next seconds (arousal, attention,
endogenous rhythms) differs between viewers and moments and is treated as noise.

\paragraph{Brain world models.}
World models learn how an environment evolves under an agent's actions~\cite{ha2018world,hafner2019planet,
hafner2020dreamer,hafner2023dreamerv3,micheli2023iris,bruce2024genie}; in their usual settings the observation is
rich and the actions strongly drive what comes next. For the brain, latent dynamical models infer neural states from
spiking activity~\cite{pandarinath2018lfads,ye2021ndt}, and NeuroWorld~\cite{neuroworld2026} rolls out fMRI with the
stimulus as the action, starting from a single observed volume. Neither transfers directly to EEG. First, the scalp
signal is a noisy mixture driven mostly by ongoing activity, so the state must be inferred from seconds of history,
not one snapshot, and the video is a weak input to strong internal dynamics. Second, the phase of ongoing
oscillations is unpredictable seconds ahead, so reconstructing the raw signal, as pixel world models reconstruct
frames, is ill-posed; band power is predictable. Third, channel and band scales differ between subjects, so one
readout cannot serve all, while a new subject offers only minutes of data. Fourth, a slow, persistent state rewards
copying the starting point, so gains must be measured against baselines with the same history; NeuroWorld's only
linear baseline is stimulus-only and its identification metric leaves the candidate pool unspecified, choices we show
inflate gains (\cref{sec:benchmark,sec:exp-ident}). EEG foundation models~\cite{jiang2024labram,wang2025cbramod,
reve2025} and fMRI foundation models~\cite{wang2026omnifmri} are evaluated mostly on classification, not forecasting,
and EEG representations align with video and language models only modestly above a time-shifted
control~\cite{platonicbrain2026}. \method{} follows from these
properties. It infers a latent state from seconds of EEG, forecasts band power, adds the video as a zero-initialized
action to dynamics pretrained on continuous EEG, and reads out through shared and subject-specific linear heads.

%% file: sections/data.tex
\section{The \dataname{} dataset}
\label{sec:data}

\begin{figure}[t]
  \centering
  \includegraphics[width=\linewidth]{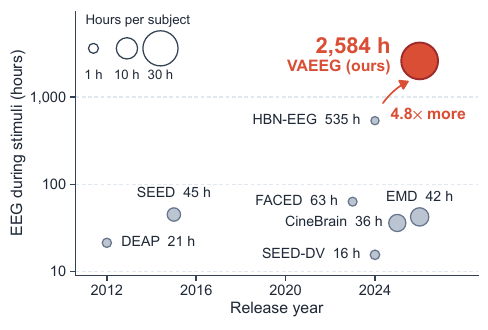}
  \caption{\textbf{Scale against public EEG datasets recorded during audio-visual stimuli.} Hours of EEG while a
  stimulus is on screen, summed over subjects; bubble area shows hours per subject.}
  \label{fig:scale}
\end{figure}

\paragraph{Corpus.}
We collected \dataname{}, 2{,}584\,h of EEG recorded from 78 subjects while short video clips were on screen, in
618{,}106 usable presentations (\cref{tab:data}). This is 4.8 times the largest public EEG dataset recorded during
audio-visual stimuli~\cite{shirazi2024hbneeg}, and at 33\,h per subject it follows each person about five times
longer than the others~\cite{koelstra2012deap,zheng2015seed,chen2023faced,liu2024eeg2video,gao2025cinebrain,gifford2026emd}
(\cref{fig:scale}). Of these hours, 2{,}122 were recorded on four public video datasets; the rest, mostly from an
earlier recording phase, on other short videos. The four video datasets, the split of the AVE clips into 3{,}312
training, 391 validation and 394 test clips, which an ancillary file lists by public clip identifier, and every
preprocessing, model and evaluation setting are public or fully specified.

\paragraph{Acquisition paradigm.}
EEG was recorded with a 32-channel wet-electrode cap (BrainCo) that streams to the presentation computer over a local
network. It has 31 scalp electrodes at standard 10--20 and 10--10 positions (\cref{fig:interp}; 30 in the earliest
recordings) and one electro-oculographic channel (IO), which we exclude. The reference electrode is at FCz and the
ground (bias) electrode at AFz. Recordings are organized in rounds of about 30\,min with self-paced breaks in between,
and subjects watched following instructions from the staff of the acquisition center.
The public clips come from four video datasets, namely AVE (10-s audio-visual event clips in 28
categories)~\cite{tian2018ave}, MUSIC-AVQA~\cite{li2022musicavqa}, the Perception
Test~\cite{patraucean2023perceptiontest} and the EEG Moments Dataset (EMD; 3-s naturalistic videos with
sound)~\cite{gifford2026emd}. Clips are shown one after another,
mostly in a fixed order, so that many subjects see each clip (each AVE test clip has a median of 33 viewers). AVE clips
start every 12\,s (10-s clip, 2-s blank). EEG was sampled at 250\,Hz, band-pass filtered (0.1--96\,Hz) with
50/60\,Hz notch filters and resampled to 256\,Hz; we keep the FCz reference.

\paragraph{Timing.}
Clip onsets are logged on the presentation computer's clock and EEG on the device's, which drifts by a median of
$-980$\,ppm; uncorrected, onsets at the end of a recording would be off by a median 1.5\,s and up to 10\,s. For each
recording we estimate the clock offset and drift rate that maximize the evoked responses to clip onsets and offsets
(fitted on odd, tested on even clips), convert every onset to the EEG clock, and exclude recordings whose fit fails
the quality criteria (validation in \cref{sup:timing}). The data bound the remaining error at our 1-s resolution.
Posterior $\alpha$ rises during the 2-s blank between clips and falls at the next onset, and in 1{,}765 of 1{,}820
recordings, 97\%, this response falls in the expected seconds.
The other 55 recordings (3\% of trials) are shifted by 1--4\,s; all models see the same data.

\begin{table}[t]
  \centering
  \caption{\textbf{\dataname{} and the subsets used.} $^\dagger$Training sessions only. $^\ddagger$Includes the
  unseen subjects' training data; the models of \cref{sec:exp-transfer} never see them. Test data never enter training.}
  \label{tab:data}
  \small
  \setlength{\tabcolsep}{4pt}
  \begin{tabular}{@{}lrrr@{}}
    \toprule
    Subset & Subjects & Hours & Use \\
    \midrule
    Full corpus & 78 & 2{,}584 & -- \\
    \quad AVE presentations & 70 & 1{,}016 & video-paired trials \\
    Continuous pool & 70$^\ddagger$ & 2{,}012 & pretraining, scaling \\
    Core subjects & 15 & 908$^\dagger$ & scaling, video, ID \\
    Unseen subjects & 24 & 516 & one-shot transfer \\
    Video fine-tuning & 65$^\ddagger$ & 630 & 226{,}979 clip viewings \\
    \bottomrule
  \end{tabular}
\end{table}

\paragraph{Features and splits.}
Each second becomes a 155-dimensional vector of log band power from a 1-s multitaper spectrum, for 31 channels
$\times$ 5 bands ($\delta$ 1--4,
$\theta$ 4--8, $\alpha$ 8--13, $\beta$ 13--30, $\gamma$ 30--45\,Hz), z-scored per recording. A second with a peak-to-peak amplitude above 500\,$\mu$V in any channel is marked bad, and
windows containing bad seconds are skipped. Continuous recordings of the core subjects are split by whole recordings
(75\% training, 10\% selection, 15\% test). Trials use the corpus's stimulus split, and a trial is the 6\,s before a clip
onset plus the 10\,s of the clip. EEG around any validation or test clip is masked out of all training data. Each clip
second is described by 24 principal components each of OpenCLIP~\cite{cherti2023openclip} and
DINOv2-L~\cite{oquab2024dinov2} frame embeddings, loudness and spectral flux, and a clip-on flag.

%% file: sections/method.tex
\section{\method}
\label{sec:method}

\begin{figure*}[t]
  \centering
  \includegraphics[width=\textwidth]{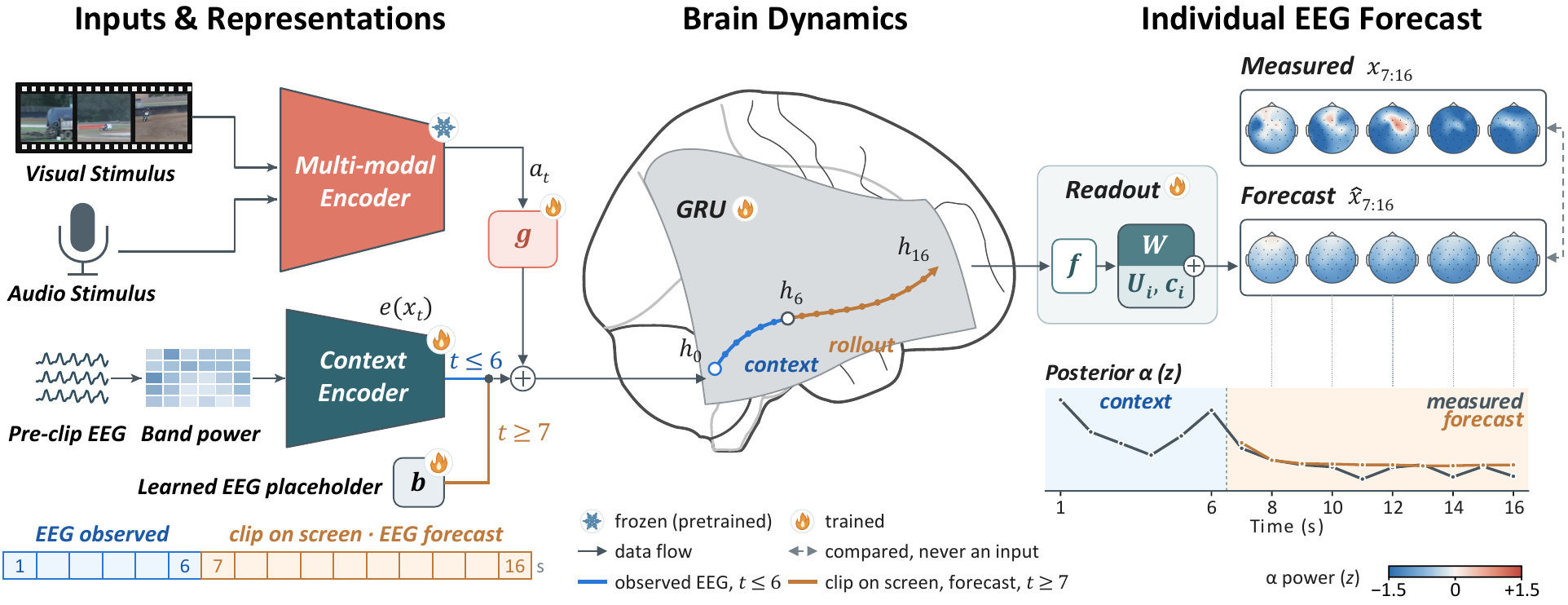}
  \caption{\textbf{\method.} One GRU step per second. In the context, the encoded band power $e(x_t)$
  updates the latent state; in the rollout a learned blank $b$ replaces it (\cref{eq:core}). The clip acts through
  the video pathway $g$ on frozen features, and a shared readout $W$ plus a per-subject head $(U_i,c_i)$ give the
  forecast. Right, a real test viewing chosen for illustration, at the 99.8th percentile of skill.}
  \label{fig:framework}
\end{figure*}

\paragraph{Task.}
Let $x_t\in\mathbb{R}^{D}$ be the EEG features of time step $t$ (band power per channel and frequency band;
\cref{sec:data}) and $a_t$ the features of the clip on screen ($a_t=0$ before onset). Given a context $x_{1:T_c}$ and
the actions $a_{1:T}$, with $T=T_c+T_f$, a model forecasts $\hat x_{T_c+1:T}$ in one free rollout. On continuous EEG without
video, the same task is posed on every window of length $T$.

\paragraph{Dynamics core (\cref{fig:framework}).}
An encoder $e$ maps each observed step to $\mathbb{R}^H$, and a GRU~\cite{cho2014gru} updates a latent brain state
$h_t\in\mathbb{R}^H$. During the rollout no EEG is observed, so the encoder output is replaced by a learned
\emph{blank} vector $b$:
\begin{equation}
  h_t=\mathrm{GRU}\big(z_t+g(a_t),\,h_{t-1}\big),\qquad
  z_t=\begin{cases} e(x_t), & t\le T_c,\\ b, & t>T_c. \end{cases}
  \label{eq:core}
\end{equation}
A trunk $f$ and a shared readout $W$ give the forecast, plus a linear head per subject $i$,
\begin{equation}
  \hat x_t = W f(h_t) + U_i f(h_t) + c_i ,
  \label{eq:readout}
\end{equation}
with $U_i,c_i$ initialized at zero. The heads absorb subject-specific channel and band scales, so the dynamics core
is shared by all subjects.

\paragraph{Video pathway.}
Each second of the clip is described by frozen features, namely OpenCLIP and DINOv2 frame embeddings, loudness and spectral
flux (\cref{sec:data}); other encoders are compared in \cref{sec:exp-interp}. The video pathway $g$ is an MLP whose
last layer is initialized at zero, so a pretrained model is unchanged at the
start of fine-tuning. Its output is added to the GRU input at every step, so the clip acts as a control input that moves
the latent state.

\paragraph{Training.}
Stage~1 pretrains the core, readout and heads on continuous EEG with the mean squared error of the free rollout
(AdamW~\cite{loshchilov2019adamw}, early stopping on the selection error). The number of steps per epoch grows in
proportion to the data, so the model sees the data equally often at every scale. Stage~2 adds the video pathway and
fine-tunes on video trials, with half of each batch drawn from continuous windows to keep the pretrained dynamics;
the pretrained parameters use a smaller learning rate than the new ones. Hyperparameters are in \cref{sup:impl}.

\paragraph{Unseen subjects.}
For a subject not seen in training, the trained model stays frozen. Because $f(h_t)$ is then fixed given the context
and the clip, the new head $(U_i,c_i)$ in \cref{eq:readout} is a ridge regression on the subject's own training
trials, shrunk toward the shared readout with the penalty chosen on their own selection trials.

\paragraph{Linear baselines.}
Every result is compared with the strongest linear model available for each subject. \emph{History ridge} maps the
$T_c\times D$ context directly to the $T_f\times D$ future, fitted per subject, pooled over subjects, or
hierarchically with per-subject weights shrunk toward the pooled ones, the linear analog of a shared core plus heads.
Adding a block of clip features gives the \emph{history + video} encoding models. For each subject, the best of the
three on their own selection data is the \emph{best linear} model. We also train a \emph{linear world model} with
\method{}'s architecture and training but every nonlinearity removed.

\paragraph{Stateless nonlinear baselines.}
To ask whether the gain needs the recurrent state or only a nonlinear network, two direct forecasters, an MLP and an
iTransformer~\cite{liu2024itransformer} that embeds each feature's history as one token, map the context, plus the
clip features on trials, to all $T_f$ steps at once, with \method{}'s inputs, per-subject heads and step budget.

%% file: sections/benchmark.tex
\section{Evaluation protocol}
\label{sec:benchmark}

Our benchmark has five tasks, namely multi-step
forecasting, use of the video through clip swapping and clip-locked $r$, clip identification, transfer to unseen
subjects and data scaling, and fixes three choices that decide the conclusions.

\paragraph{Baselines with EEG history.}
Every task reports the context mean, history-only ridge and history + video ridge, each per subject, pooled and
hierarchical; linear models are hard to beat on brain data~\cite{schulz2020different,he2020deep} and time
series~\cite{zeng2023transformers}. A stimulus-only linear model, the usual encoding baseline~\cite{neuroworld2026,gifford2025algonauts},
reaches a per-step correlation of only 0.04 on our trials, against 0.23 once EEG history is added; comparing with it
would overstate any forecaster's gain several-fold. On trials, hierarchical ridge beats per-subject
ridge by 0.008 skill, so comparing with per-subject ridge alone would double the apparent gain of the world model.

\paragraph{Metrics.}
The headline metric is the forecast correlation $r$, as in brain encoding~\cite{neuroworld2026,gifford2025algonauts}.
It is the Pearson $r$ across test windows for each forecast step and each feature, averaged over features and steps. Because
$r$ ignores amplitude, we also report \emph{skill}, the squared error relative to the context-mean forecast
$\bar x^{\,\mathrm{ctx}}$,
\begin{equation}
  \mathrm{skill}=1-\frac{\sum_{t=T_c+1}^{T}\sum_{d=1}^{D}(\hat x_{t,d}-x_{t,d})^2}{\sum_{t=T_c+1}^{T}\sum_{d=1}^{D}(\bar x^{\,\mathrm{ctx}}_{d}-x_{t,d})^2},
\end{equation}
which rewards calibrated amplitude. Pre-registered decisions used skill; every comparison that is significant under
either metric points the same way under both, except zero-shot transfer (\cref{sec:exp-transfer}). To make
differences tangible we also express them in time. The \emph{horizon gained} is how many seconds earlier the baseline's
per-step $r$ falls to the level the model reaches at the last step (linear interpolation between steps, per
subject).
Trial results are averaged within each test clip and then across clips, so frequently shown clips do not dominate.

\paragraph{Persistence-controlled identification.}
To ask whether a forecast knows \emph{what} is being watched, we hold the real context fixed and forecast once for
each test clip; the true clip should give the forecast closest to the recorded EEG. A model
that ignores the video scores exactly chance. We also report results under the evaluation
protocol of comparable work~\cite{neuroworld2026} (\cref{sec:exp-ident}).

\paragraph{Statistics and pre-registration.}
The unit of inference is the subject. We use exact sign-flip tests over the 15 core subjects, 100{,}000 random flips over
the 24 unseen ones, and for trials also a sign-flip test over the 394 test clips. Every hypothesis, its decision rule
(mean difference $>0$, $p<0.05$, at least 11/15 or 16/24 subjects positive) and every use of a test set was written
down before the run~\cite{nosek2018preregistration}; second attempts use $p<0.025$. All of them, including the failed ones, are in \cref{sup:prereg}.

%% file: sections/experiments.tex
\section{Experiments}
\label{sec:experiments}

We use $T_c=6$, $T_f=10$ steps of 1\,s, $D=155$ and $H=512$; the remaining hyperparameters are in \cref{sup:impl}.
All results are on held-out recordings of the 15 core subjects for continuous EEG and on their 10{,}970
viewings of the 394 test clips for trials. Every comparison below was tested per subject; the intervals, subject
counts and $p$-values, and which tests were pre-registered, are listed in \cref{sup:stats,sup:prereg}.

\subsection{Forecasting continuous EEG}
\label{sec:exp-scaling}

\begin{figure}[t]
  \centering
  \includegraphics[width=\columnwidth]{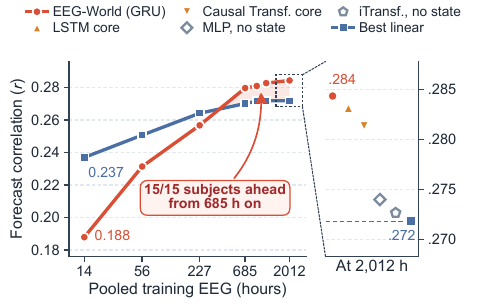}
  \caption{\textbf{Scaling on continuous EEG.} Forecast correlation on held-out recordings of the 15 core
  subjects against pooled training EEG; shaded, EEG-World's lead. Right, all cores and the stateless forecasters at
  2{,}012\,h.}
  \label{fig:scaling}
\end{figure}

As in scaling studies~\cite{kaplan2020scaling,hoffmann2022chinchilla}, we pretrain \method{} without video on nested
pools of continuous EEG and compare it at every size with the best linear history model fitted on the same data
(\cref{fig:scaling}). The ranking reverses with scale. With tens of hours the linear model is clearly better; the world
model overtakes it between 227 and 685 pooled hours and from then on wins in every subject. The nonlinear part of EEG
dynamics thus becomes learnable only once hundreds of hours are pooled across people, which explains why linear
forecasters are hard to beat on typical EEG datasets of a few hours per person.

\paragraph{The linear model saturates; the world model does not.}
Beyond the crossover the two models diverge. The best linear model levels off after a few hundred hours, so more data
add nothing that a linear map from the recent past can use, while the world model keeps improving up to 2{,}012\,h.
About half of this late improvement comes from the larger training budget and half from the additional data (\cref{sup:scaling}).

\paragraph{What matters is a state carried through time.}
The same architecture with every nonlinearity removed falls below closed-form ridge, so the gain is not an artifact of
training. An MLP and an iTransformer~\cite{liu2024itransformer} that map the
same context directly to the future, with the same heads and budget, barely improve on the linear model, so
nonlinearity alone is not enough. LSTM and
causal-Transformer cores, which like the GRU unroll a latent sequence step by step, gain almost as much as the GRU
(\cref{fig:scaling}, right). What the linear models miss is therefore how the brain's state evolves, not a richer map
from past to future. The GRU is both the best and the smallest core, so at this scale a compact recurrent state matters
more than capacity.

\subsection{Forecasting with the video as action}
\label{sec:exp-video}

\begin{figure}[!tb]
  \centering
  \includegraphics[width=\columnwidth]{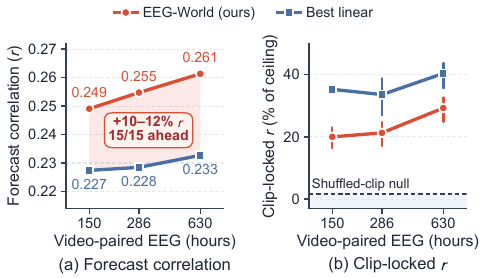}
  \caption{\textbf{Video trials.} Held-out clips of the 15 core subjects against the amount of video-paired
  training EEG. (a)~Forecast correlation. (b)~Clip-locked correlation (SLv) between the clip-driven part of each forecast
  and the mean EEG of the other viewers, in percent of the noise ceiling; shaded, below $+2$\,SD of a
  shuffled-clip null.}
  \label{fig:video}
\end{figure}

\begin{table}[t]
  \centering
  \caption{\textbf{Video trials.} The clip-onset model is fine-tuned on the same trials but told only when each
  clip starts; the no-state models are the direct forecasters of \cref{sec:method}; the swap cost is the correlation
  lost when the clip is replaced by another test clip.}
  \label{tab:video}
  \small
  \setlength{\tabcolsep}{4pt}
  \begin{tabular}{@{}lccc@{}}
    \toprule
    Model (15 subjects, test clips) & $r$ & skill & swap cost ($r$) \\
    \midrule
    Context mean & .180 & 0 & -- \\
    Linear, EEG only & .225 & .1909 & -- \\
    Linear, EEG + video & .233 & .1928 & .012 \\
    MLP, no state & .242 & .1948 & -- \\
    iTransformer, no state & .248 & .1982 & -- \\
    \method{}, EEG + clip onset & .255 & \textbf{.2034} & -- \\
    \method{}, EEG + video & \textbf{.261} & .2030 & \textbf{.026} \\
    \bottomrule
  \end{tabular}
\end{table}

On held-out clips \method{} improves on the strongest history + video encoding model by 12\% in $r$
(\cref{tab:video}). The advantage is small in the first second and grows with the horizon. To be as accurate at its
last step as \method{}, the linear model would have to stop 3.3\,s earlier. The first second after the context is
largely fixed by the immediate past, which any forecaster can copy; the world model's value lies in carrying the state
further ahead. Stateless nonlinear forecasters help more on trials than on continuous EEG but stay well below
\method{}, so the recurrent state matters with video as it does without.

\paragraph{Individual trajectories versus the average response.}
More video-paired data help the world model more than the linear model, and its clip-locked response grows with them
(\cref{fig:video}). Yet the linear encoding model follows the mean response across viewers more closely
(\cref{fig:video}b), while the world model forecasts each viewer more accurately (\cref{fig:video}a). The two targets
differ, because averaging over viewers removes the state-dependent part of the signal that the world model predicts. A
benchmark that scores models only against averaged responses therefore favors encoding models and cannot see what a
forecaster of individual brains adds.

\subsection{Transfer to unseen subjects}
\label{sec:exp-transfer}

\begin{figure*}[t]
  \centering
  \includegraphics[width=\textwidth]{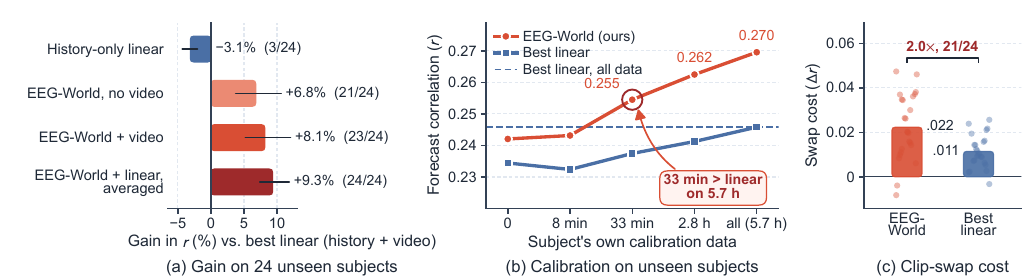}
  \caption{\textbf{Unseen subjects.} 24 subjects never used in development; frozen models with a
  per-subject linear head. (a)~Gain over the best history + video linear model, in percent of its $r$, with the subjects ahead.
  (b)~A model retrained without these subjects, against each subject's own calibration data; dashed, the best linear
  model fitted on all of it. (c)~Correlation lost when the clip is swapped; dots, subjects.}
  \label{fig:transfer}
\end{figure*}

\begin{figure*}[b]
  \centering
  \includegraphics[width=\textwidth]{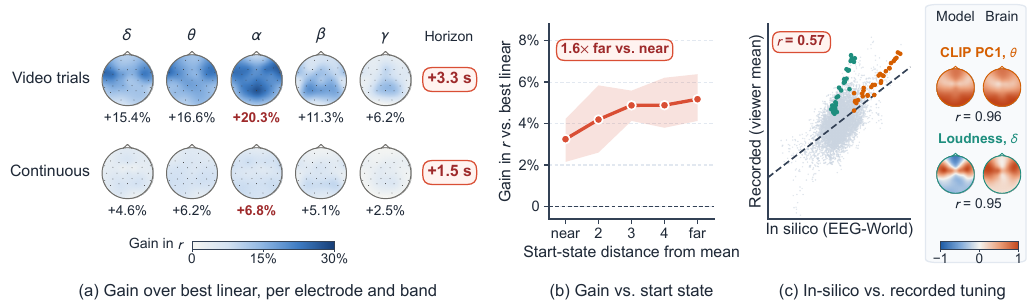}
  \caption{\textbf{Where, when and what.} (a)~Gain over the best linear model at each electrode and band, in percent of its $r$,
  and the horizon gained. (b)~Gain, in percent, against the distance of the starting state from the subject's mean. (c)~In-silico
  tuning, the forecast change when one input feature is raised, against the tuning of recorded EEG across viewers,
  for every feature and electrode--band cell; right, maps of two features. Panel (c) uses validation data only.}
  \label{fig:interp}
\end{figure*}

\begin{figure*}[t]
  \centering
  \includegraphics[width=\textwidth]{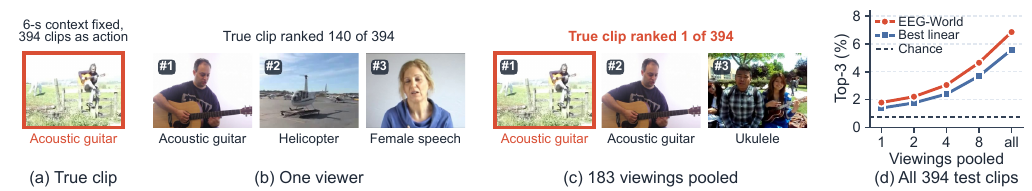}
  \caption{\textbf{Identifying the clip from forecasts.} Each viewing keeps its real context and is forecast with each
  of the 394 test clips as the action; candidates are ranked by fit to the recorded EEG. (a)~An example test clip,
  shown by one frame and its event label. (b)~The viewing with the median rank of the true clip. (c)~Mean over all
  183 viewings of this frequently repeated clip. (d)~All test clips, pooling random subsets of viewings. Post hoc.}
  \label{fig:retrieval}
\end{figure*}

We tested 24 subjects never used during development, once, with the analysis fixed in advance; both the subjects and
the clips are new to the model. With the video model frozen and only a linear head fitted to each subject, \method{}
beats the best history + video linear model in 23 of the 24 (\cref{fig:transfer}a); a model retrained on the other 41
subjects replicates the result (\cref{fig:transfer}b).

\paragraph{What transfers is the dynamics.}
A world model that never saw video transfers almost as well as the video model, although the video model does use the
clip (\cref{fig:transfer}c). What carries over to a new brain is mainly the dynamics learned from continuous EEG; the
stimulus mapping learned on the training cohort adds little for someone new. Averaging world-model and linear
forecasts beats both, so the two make partly independent errors. On two other EEG devices a frozen model with only a
new head merely matches the linear models (\cref{sup:devices}), so the dynamics transfer across people within a recording setup
but not yet across devices.

\paragraph{Calibration.}
With only \textbf{33\,min} of a new subject's data for the head, the retrained model reaches $r=0.255$, above the
$0.246$ of a
linear model fitted on all of that subject's data, a median of \textbf{5.7\,h} or about ten times more
(\cref{fig:transfer}b). Without calibration it already correlates better
than the pooled linear model but is not more accurate in squared error, because the shared readout gets the shape of the
trajectory right and the scale wrong. People differ mainly in channel and band scales, which a linear head absorbs, and
much less in their dynamics, which the shared core captures.

\subsection{What the world model learns}
\label{sec:exp-interp}

\paragraph{The gain covers the whole scalp, most in $\alpha$.}
It is positive at every electrode and band, on continuous EEG and on video trials (\cref{fig:interp}a). A model that had learned a local response, such as occipital $\alpha$ to visual
input, would give a localized map; a gain spread over the whole scalp points to a global brain state. On video trials
it is largest in $\alpha$ ($+20\%$), the rhythm most closely tied to visual attention and arousal, and about three times as large
as on continuous EEG.

\paragraph{The gain is largest far from the brain's usual state.}
A linear forecaster can only let each feature relax toward its mean at a fixed rate. If the brain's return depends on
where it starts, a nonlinear model should gain most when the starting state is unusual. Indeed, the gain grows
steadily with the distance of the starting state from the subject's mean, and the farthest fifth of starting states
gains $1.6\times$ as much as the nearest in $r$ ($+5.2\%$ against $+3.2\%$; \cref{fig:interp}b) and $3.8\times$ as
much in skill. The advantage is concentrated where linear relaxation
is least adequate, in states far from equilibrium. The forecasts are smooth but not shrunk toward the mean (\cref{sup:calibration}).

\paragraph{In-silico tuning matches recorded EEG.}
We use \method{} as a virtual subject. Keeping the real context, we raise one video or audio feature for the whole clip
and record the change in the forecast. On held-out validation clips we measure the same tuning in recorded EEG from the
mean response of all viewers of each clip. The two agree across features and channels (\cref{fig:interp}c); for the
strongest feature, the first CLIP component, which separates plucked-string instruments from vehicles, the
$\theta$-band maps nearly coincide. The model was trained only to forecast single viewings and never saw an averaged
response, so this agreement is not built in. Forecasting individual viewers forces the video pathway to learn the
stimulus--response relations that population averages reveal.

\begin{table}[t]
  \centering
  \caption{\textbf{Ranking clip features.} The video pathway is fed one feature set at a time, 24
  principal components each, three seeds. SLv as in \cref{fig:video}b, without noise-ceiling normalization.}
  \label{tab:encoders}
  \footnotesize
  \setlength{\tabcolsep}{4pt}
  \begin{tabular}{@{}llcc@{}}
    \toprule
    Clip features & Modality & SLv & $r$ \\
    \midrule
    V-JEPA 2~\cite{assran2025vjepa2} & Video & \textbf{.188} & \textbf{.261} \\
    CLIP + DINOv2 (main visual input) & Image & .139 & .260 \\
    CLIP~\cite{cherti2023openclip} & Image & .133 & \textbf{.261} \\
    DFN~\cite{fang2024dfn} & Image & .126 & .255 \\
    VideoMAE~\cite{tong2022videomae} & Video & .115 & .255 \\
    MAE~\cite{he2022mae} & Image & .110 & .255 \\
    DINOv2~\cite{oquab2024dinov2} & Image & .109 & .253 \\
    \midrule
    Log-mel spectrogram & Audio & .062 & .249 \\
    Loudness + spectral flux & Audio & .061 & .259 \\
    \midrule
    Clip onset only & -- & 0 & .253 \\
    \bottomrule
  \end{tabular}
\end{table}

\paragraph{Ranking vision encoders.}
Because the clip enters \method{} only through its features, swapping the encoder turns the model into a test of
vision models (\cref{tab:encoders}). The forecast correlation hardly changes across encoders, since it is dominated by
the brain's own dynamics; the clip-locked response, which isolates the clip-driven part of the forecast, separates them
clearly. V-JEPA 2~\cite{assran2025vjepa2}, trained to predict masked parts of videos in its latent space, drives this
response best, 35\% above our CLIP + DINOv2 input. Temporal input alone does not explain this, because VideoMAE, also a
video model but trained to reconstruct pixels, ranks with the image models. The objective of predicting video in a latent
space thus appears to matter more than seeing motion. The main results keep the pre-registered CLIP + DINOv2 input.

\subsection{Stimulus identification}
\label{sec:exp-ident}

Under NeuroWorld's trajectory-matching protocol~\cite{neuroworld2026}, \method{} finds the right trial among the
subject's test viewings far above chance and slightly more often than the linear model. But a ``model'' that ignores
the video and repeats the context mean already reaches $12.5\times$ chance, and widening the candidate pool to all
viewings moves the same forecasts from $16\times$ to $145\times$ chance. The protocol mainly measures how persistent
the starting state is and how the pool is drawn. Under our persistence-controlled protocol, where ignoring the video
scores exactly chance, both \method{} and the linear model identify the clip above chance but are not separated, because a
single viewing carries little clip-specific information in band power. This information adds up across viewers
(\cref{fig:retrieval}). Averaging each candidate's score over all viewings of a test clip, each forecast from its own
context, raises top-1 identification from 0.8\% to 3.8\% of the 394 clips (chance 0.25\%; linear model 3.3\%), and
the top candidate shares the true clip's category in 13\% of clips (chance 4\%). This pooled analysis is post hoc.

%% file: sections/discussion.tex
\section{Discussion and limitations}
\label{sec:discussion}

\paragraph{The brain is the world.}
In every analysis the strongest predictor of the next seconds of EEG was the brain's own state, and what transferred
to unseen subjects was the dynamics learned from continuous EEG. A brain-based benchmark for vision that ignores these
dynamics attributes to the stimulus what belongs to the brain; history-aware baselines and persistence-controlled
metrics are cheap and should be standard. Because the advantage grows with training EEG, forecasting is a natural
pretraining objective for EEG foundation models.

\paragraph{Limitations.}
We forecast one representation (per-second log band power); finer time scales and source-space targets remain open.
Video helped the trained cohort but added no net gain on unseen subjects. The core subjects' test sets were used in
several pre-registered analyses, all listed in \cref{sup:prereg}; the confirmatory transfer result uses 24 subjects tested
once. Clips were shown in a largely fixed order, so carry-over from the previous clip is controlled only through the
swap analyses.

\paragraph{Ethics.}
The study was approved by the Science and Technology Ethics Committee of the Southern University of Science and Technology (approval no.~20260399). All participants gave written informed consent and were compensated.

%% file: sections/llm.tex
\section*{LLM Usage Statement}
LLM-based assistants were used under the authors' direction. Claude (Anthropic) was used to write and debug analysis
code, draft figures, and polish the text of this paper. ChatGPT and Codex (OpenAI) and Gemini (Google) were used to review
the text. All hypotheses, decision rules and results were reviewed by the authors, who take full responsibility for
the content.

%% file: sections/appendix.tex
\section{Implementation details}
\label{sup:impl}

\paragraph{Sizes.} Context $T_c=6$ and rollout $T_f=10$ steps of 1\,s (16-s windows); $D=155$ features (31 channels
$\times$ 5 bands of log power); $H=512$. The encoder $e$ is linear, LayerNorm, GELU and dropout 0.1; the trunk $f$ is
linear and GELU; the video pathway $g$ is a two-layer MLP whose last layer is initialized at zero. The linear world
model replaces the GRU by a linear recurrence and drops every GELU. The GRU core has 1.6M parameters and the shared
model 2.0M (2.3M with the video pathway); each subject head adds 80k, so the 2{,}012-h model with its 70 heads has
7.6M.

\paragraph{Optimization.} Both stages minimize the mean squared error over the $T_f$ rollout steps only; gradients
flow back through all 16 GRU steps, including the context, and no observed EEG enters during the rollout (no teacher
forcing). Stage~1: AdamW~\cite{loshchilov2019adamw}, learning rate $10^{-3}$, weight decay 0.01,
batch 64, warm-up and cosine decay over 80 epochs, early stopping on the selection error; 250 steps per epoch at
908\,h and 555 at 2{,}012\,h. Stage~2: video trials of 65 subjects (226{,}979 viewings, 630\,h), half of each batch
from continuous windows (chosen on validation); learning rate $3\!\times\!10^{-4}$ for the pretrained parameters
(selected from four candidate rates on validation clips) and $10^{-3}$ for the video pathway and new heads.

\paragraph{Baselines.} Selection data are 10\% of each subject's recordings for continuous EEG and the viewings of the
391 validation clips for video trials. All models, linear or not, choose their hyperparameters on these data, and the
trained ones also their stopping epoch; none is refitted on them. Ridge penalties are chosen per subject from $10^{-4}$ to $10^{5}$
times the mean eigenvalue of the design matrix, extended to $10^{7}$ for the shrinkage of hierarchical ridge toward
the pooled weights; for history + video ridge, the scale of the video block is chosen from 1, 0.3, 0.1, 0.03 and 0.01.
The stateless baselines, a three-layer MLP and an iTransformer~\cite{liu2024itransformer}, choose width, depth and
learning rate in the same way.

\paragraph{Models for unseen subjects.} The 2{,}012-h continuous pool and the 65-subject video set include the
training data of the 24 unseen subjects, so the models of \cref{sec:exp-transfer} are trained without them. \Cref{fig:transfer}a uses the frozen
video model fine-tuned on 54{,}133 trials before these subjects were used; \cref{fig:transfer}b and the zero-shot test use a model
fine-tuned from the same pretrained core on the other 41 subjects of the video set. The pooled and hierarchical linear
baselines are refitted on the same subjects as each model, so no comparison involves data of the 24 subjects beyond
their own calibration data.

\section{Data}
\label{sup:data}

\paragraph{Public assets.} Of the 2{,}584\,h of EEG, 2{,}122\,h were recorded on four public video datasets:
AVE~\cite{tian2018ave} (4{,}097 clips of 10\,s in 28 event categories; 1{,}016\,h), MUSIC-AVQA~\cite{li2022musicavqa}
(536\,h), the Perception Test~\cite{patraucean2023perceptiontest} (510\,h) and the EEG Moments
Dataset~\cite{gifford2026emd} (60\,h). The remaining 462\,h, mostly from an earlier recording phase, were recorded on
other short videos. Only AVE clips enter the video experiments; the other three public datasets contribute continuous
EEG. The ancillary file \texttt{ave\_clip\_split.csv} lists the public
identifier of every AVE clip and whether it is a training (3{,}312), validation (391) or test (394) clip. All stimulus
features come from frozen public encoders.

\paragraph{In-house assets.} The EEG recordings are human-subject data collected under an ethics-approved protocol
with written informed consent. All analyses identify participants by pseudonymous codes only.

\paragraph{EEG preprocessing.} The cap samples at 250\,Hz with the reference at FCz; the corpus preprocessing
band-passes at 0.1--96\,Hz, applies 50 and 60\,Hz notch filters and resamples to 256\,Hz. Clip onsets are mapped to the EEG
clock as described below. Each recording is cut into 1-s bins on the corrected EEG time axis, starting 6\,s before
its first clip onset. For every bin and channel, the power spectrum is a multitaper estimate with discrete prolate
spheroidal tapers of time--half-bandwidth product 2, keeping tapers with spectral concentration above 0.9, after
removing the mean and weighting the tapers by their eigenvalues. Band power is the mean spectral power over the
frequency bins of $\delta$ 1--4, $\theta$ 4--8, $\alpha$ 8--13, $\beta$ 13--30 and $\gamma$ 30--45\,Hz, and the
feature is its base-10 logarithm, giving $31\times5=155$ features per second. A bin is marked bad when any channel's
peak-to-peak amplitude exceeds 500\,$\mu$V; features are z-scored per recording over its good bins, and windows that
contain a bad bin are skipped.

\paragraph{Stimulus features.} For each second of a clip, OpenCLIP~\cite{cherti2023openclip} and
DINOv2-L~\cite{oquab2024dinov2} image embeddings are averaged over four frames; each is reduced to 24 principal
components fitted on the training clips and scaled to unit variance there. The sound track, converted to 16-kHz mono,
gives the loudness of each second as its RMS level in dB, floored at 16-bit silence, and its spectral flux as the mean
positive change of log short-time power with a 10-ms hop; both are z-scored on the training clips. A flag marks the
seconds in which a clip is on screen, and all features are zero before onset. The encoders compared in \cref{tab:encoders} are used in the same way, with 16 frames per second for the video models.

\paragraph{Clock correction.}
\label{sup:timing}
The device clock differs from the presentation computer's by a median of $-980$\,ppm (IQR $-1{,}430$ to $-760$).
Left uncorrected, onsets at the end of a recording (median 31\,min) would be off by a median 1.5\,s, 4.5\,s at the
90th percentile and up to 10\,s. The clock offset and drift rate of each recording are fitted with a matched filter on
1--30\,Hz EEG around clip onsets and offsets, on odd clips, and tested on even clips. Where the acquisition software
also logged device sample counts, the two estimates agree (median difference 1\,ppm, $r=0.994$, 501 recordings).
The data-driven timing check in \cref{sec:data} (posterior $\alpha$ at the 2-s blank, 1{,}765 of 1{,}820
recordings in the expected seconds) uses the training clips of recordings with at least 40 viewings. Without the
correction the video gain of a linear model nearly halves. The constant offset (median 51\,ms) includes
the mean display and transmission delay; the absolute latency was not measured with a photodiode.

\section{Viewers' states persist through a clip}
\label{sup:state}

\begin{figure*}[t]
  \centering
  \includegraphics[width=\textwidth]{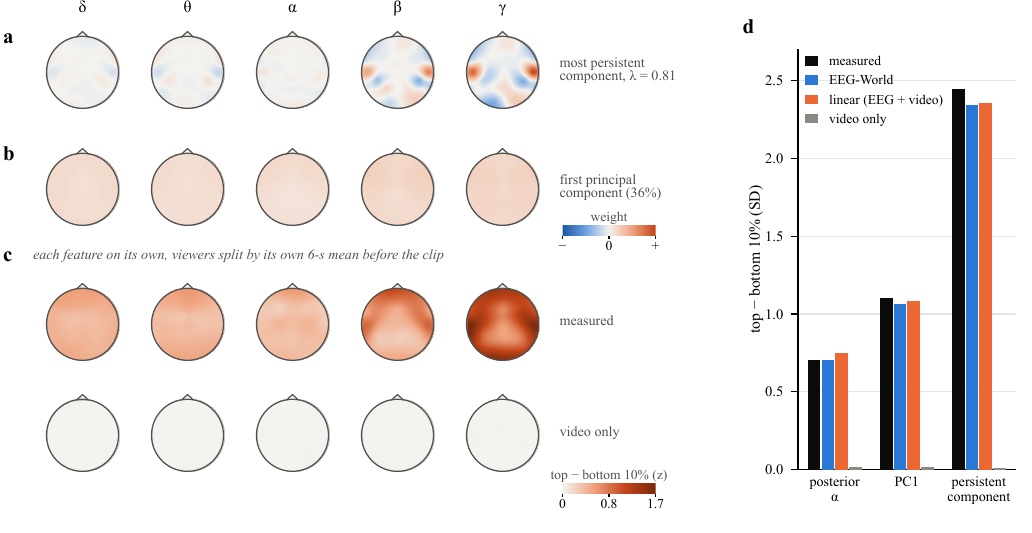}
  \caption{\textbf{Viewers who start a clip in different states stay different.} All 10{,}970 test viewings of the
  15 core subjects. (a)~Weights of the most persistent component of the 155 features, learned from training viewings
  only: the direction whose mean over the 6\,s before a clip correlates best with its mean over seconds 2--10 of the
  clip, after removing each clip's mean response over the other viewers ($\lambda$ is that correlation). (b)~The first
  principal component of the pre-clip means (36\% of their variance). (c)~Each feature on its own: viewers in the top
  and bottom 10\% of the feature's pre-clip mean, their difference over seconds 2--10 of the clip, measured and
  predicted from the video alone (the mean response of the other 14 subjects to the same clip). (d)~The same
  difference for three state variables.}
  \label{fig:state}
\end{figure*}

\Cref{fig:teaser} rests on the observation that viewers begin the same clip in different states and that the
difference outlasts the clip's onset. \Cref{fig:state} tests this on every test viewing. We split the viewings of each
clip by a feature's mean over the 6\,s before onset and follow the top and bottom 10\% through the clip. For 141 of
the 155 features the two groups still differ over seconds 2--10 of the clip in all 15 subjects, and for all 155 in at
least 13 (\cref{fig:state}c). The prediction from the video alone, the mean response of the other subjects to the same
clip, carries almost none of this difference (median 0.3\% of the measured one), so a stimulus-only encoding model
cannot express it.

The most persistent direction keeps a correlation of $\lambda=0.81$ between its pre-clip and in-clip means. Its
weights lie on $\beta$ and $\gamma$ power at the temporal electrodes (T7, T8; \cref{fig:state}a), a pattern typical of
scalp muscle activity, so part of what persists is likely muscle tone rather than neural activity. The first principal
component weights all features about equally and keeps a correlation of 0.38; posterior $\alpha$ keeps only 0.25.
Along all three
directions, \method{} and the linear model with EEG history and video keep the groups as far apart as they are
measured (2.3 and 2.4\,SD against 2.4\,SD for the persistent component), while the video-only prediction does not
(0.01\,SD; \cref{fig:state}d).

Two conclusions follow. A large part of each viewer's response is carried over from before the clip, which is why our
baselines see the same EEG history rather than the stimulus alone. And carrying a state forward does not by itself
separate the world model from the linear model, since any history-aware forecaster does it; the world model's
advantage lies in how the state evolves, most of all when it starts far from the subject's usual state (\cref{fig:interp}b).

\section{What the gain is made of}
\label{sup:ablation}
\label{sup:onset}

\begin{table}[h]
  \centering
  \caption{\textbf{Ablations} on the core subjects' test viewings, one change at a time, three seeds
  each.}
  \label{tab:ablation}
  \footnotesize
  \setlength{\tabcolsep}{4pt}
  \begin{tabular}{@{}lccc@{}}
    \toprule
    Variant (video trials) & $r$ & $\Delta r$ [95\% CI] & full ahead \\
    \midrule
    \method{} (full) & .260 & -- & -- \\
    no per-subject heads & .240 & $-.020$ [$-.023$, $-.017$] & $15/15$ \\
    3-s context (not 6\,s) & .249 & $-.011$ [$-.013$, $-.009$] & $15/15$ \\
    no pretraining & .251 & $-.009$ [$-.011$, $-.007$] & $14/15$ \\
    clip onset only & .253 & $-.007$ [$-.012$, $-.002$] & $10/15$ \\
    no continuous windows & .258 & $-.0016$ [$-.0030$, $-.0004$] & $11/15$ \\
    blank $b=0$ & .259 & $-.0006$ [$-.0012$, $+.0002$] & $11/15$ \\
    \bottomrule
  \end{tabular}
\end{table}

Removing one component at a time (\cref{tab:ablation}) shows that the per-subject heads matter most ($-0.020$ in $r$,
$15/15$; skill $-0.010$): a single readout cannot fit everyone's channel and band scales. Reading only the last 3 of
the 6 context seconds costs $0.011$ ($15/15$), so the state needs several seconds of history, and training on video
trials alone, without pretraining on continuous EEG, costs $0.009$ ($14/15$). Each of the three costs more than the
clip's content ($0.007$). Continuous windows in each fine-tuning batch help a little ($0.0016$, $11/15$), and fixing
the blank at $b=0$ in both stages changes $r$ by only $-0.0006$, with an interval that includes zero, so the learned
blank is not essential.

The clip reaches the model in two ways, as the moment it starts and as its content. Telling the model only when the
clip starts already yields nearly all of the gain in skill (\cref{tab:video}): much of what happens at clip
onset is a generic shift of brain state, shared by all clips. The content adds a smaller improvement in $r$, about as
much as it adds to the linear model. It is used, since replacing the clip costs \method{} twice as much as it costs
the linear model, yet even without it \method{} beats the linear EEG + video model ($+0.021$ in $r$, $15/15$; skill
$+0.009$, $14/15$; means of three seeds, descriptive). The world model's advantage lies in the dynamics that the
stimulus perturbs, not in a better stimulus encoding.

\section{Where the clip's features act}
\label{sup:av}

\begin{figure*}[t]
  \centering
  \includegraphics[width=\textwidth]{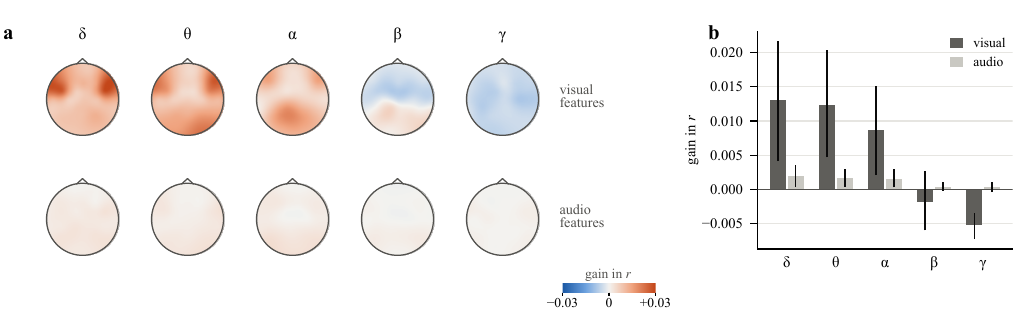}
  \caption{\textbf{Visual and audio features.} Gain in forecast $r$ on the test clips from each kind of feature: the
  main effect in a $2\times2$ design of models retrained with visual features, audio features, both or neither (three
  seeds each), averaged over the 10 forecast steps. (a)~Subject means at each electrode and band, on one scale.
  (b)~Band means over all electrodes with bootstrap 95\% intervals over the 15 subjects.}
  \label{fig:av}
\end{figure*}

To see which features the video pathway uses, we retrained it with only the visual features, only the audio
features or neither, and measured the gain each kind of feature brings (\cref{fig:av}). The pathway relies on the
visual features. Their largest gains are in $\delta$ and $\theta$ at the lateral frontal electrodes (F7, F8, FT9,
FT10; up to $+0.03$ in $r$), where horizontal eye movements show, and in $\theta$ and $\alpha$ over
parieto-occipital electrodes ($+0.008$ in $r$ there, $11/15$, pre-registered). In $\gamma$ they slightly lower $r$
($-0.005$). Part of the visual effect is therefore likely oculomotor, since a clip's content predicts how viewers move
their eyes as well as how their visual cortex responds, and the parieto-occipital gain is not significantly larger
than the temporal one ($p=0.07$). The audio features add little ($+0.001$ over temporal sites, $12/15$, pre-registered, no
larger than elsewhere) and nothing once the visual features are present. Swapping only the visual features costs
$0.022$--$0.029$ in $r$ in every region; swapping only the audio features, ${<}0.001$.

\section{Scale, compute and architecture}
\label{sup:scaling}

\begin{table}[h]
  \centering
  \caption{\textbf{Continuous EEG, no video.} Held-out sessions of the 15 core subjects, seed means;
  parameters of the core only. $^\ast$\method{}'s architecture and training without nonlinearity.}
  \label{tab:scaling}
  \footnotesize
  \setlength{\tabcolsep}{3pt}
  \begin{tabular}{@{}lrcccc@{}}
    \toprule
    & & \multicolumn{2}{c}{908\,h} & \multicolumn{2}{c}{2{,}012\,h} \\
    \cmidrule(lr){3-4}\cmidrule(l){5-6}
    Model & Params & $r$ & skill & $r$ & skill \\
    \midrule
    Best linear (history ridge) & -- & .2716 & .1745 & .2718 & .1746 \\
    Linear world model$^\ast$ & -- & .2572 & .1655 & -- & -- \\
    \midrule
    \method{} (GRU) & 1.6M & \textbf{.2809} & \textbf{.1787} & .2844 & .1804 \\
    \quad + JEPA latent loss & 1.6M & -- & -- & \textbf{.2859} & \textbf{.1813} \\
    \quad 908\,h data, 2{,}012\,h budget & 1.6M & .2824 & .1795 & -- & -- \\
    LSTM core & 2.1M & .2784 & .1773 & .2831 & .1797 \\
    Causal Transformer core & 6.3M & .2788 & .1775 & .2813 & .1787 \\
    \midrule
    \multicolumn{6}{@{}l}{\emph{Nonlinear, no state (direct 10-s forecast)}} \\
    MLP & 5.7M & -- & -- & .2740 & .1745 \\
    iTransformer & 3.6M & -- & -- & .2727 & .1740 \\
    \bottomrule
  \end{tabular}
\end{table}

About half of the gain from 908 to 2{,}012\,h comes from compute, as the 908-h data trained with the 2{,}012-h step
budget reach $r=0.282$ and the larger pool $0.284$ (\cref{tab:scaling}). The LSTM and causal Transformer cores (2 or 4 layers,
learning rate tuned separately; up to 4$\times$ the core parameters) beat the best linear model in $15/15$ subjects at
908 and 2{,}012\,h, so the gain does not depend on the GRU; the GRU is the best and the smallest, and the Transformer
is $0.003$ lower in $r$ at 2{,}012\,h ($0/15$). An auxiliary loss on the latent state of an EMA target
encoder~\cite{lecun2022path,assran2025vjepa2} adds a small, consistent gain (skill $15/15$).

\section{Calibration: smooth forecasts are not a collapse to the mean}
\label{sup:calibration}

The forecasts are smoother than the EEG, as forecasts of an uncertain future must be. To check that they are not
simply shrunk toward the mean, we regressed the measured EEG on the forecast, per subject, step and feature. On
continuous EEG the slope is 0.93--0.99 at every step, and the pre-registered criterion (slope within $[0.8,1.25]$ at
all 10 steps in at least 13 of 15 subjects, on validation and test data) holds ($14/15$ and $15/15$). On video
trials the slope falls from 0.93 at the first second to 0.81 at the tenth, so the criterion holds in only $6/15$: late
forecasts slightly \emph{overshoot}, the opposite of a collapse, and the best linear model does the same (0.95 to
0.79). In both settings the forecast variance is close to $r^2$ times the measured variance, as for a calibrated
forecast.

\section{Identification}
\label{sup:ident}

\begin{figure}[h]
  \centering
  \includegraphics[width=\columnwidth]{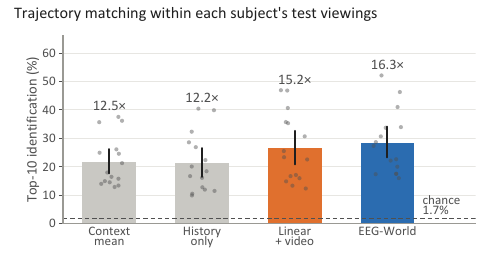}
  \caption{\textbf{Trajectory matching}~\cite{neuroworld2026} within each subject's test viewings: top-10
  accuracy per subject with mean and 95\% CI, in multiples of chance. The context mean ignores the video.}
  \label{fig:ident}
\end{figure}

Under the trajectory-matching protocol (\cref{fig:ident}) \method{} reaches 28.5\% top-10 ($16\times$ chance) and the
best linear model 26.5\% ($+1.9$ points, $11/15$, $p=0.04$), but the context mean already reaches $12.5\times$ chance.
With all 10{,}970 viewings of the 15 subjects as candidates, the same forecasts reach $145\times$ chance (context mean
$90\times$). Under the persistence-controlled protocol, \method{} identifies the clip at 5.5\% ($2.2\times$ chance,
$15/15$ above chance) and the linear model at 4.7\% (subject-level $p=0.15$, pre-registered, failed; clip level
$p=0.024$): single viewings carry little clip-specific signal in band power. The context mean scores so well under
trajectory matching because of the persistence shown in \cref{sup:state}, which makes a viewing's own past resemble
its future whatever the clip.

\section{Other EEG devices}
\label{sup:devices}

Reached with only a new linear head, the frozen model is on par with the best linear history model on two other
devices: $+0.004$ in $r$ [$-0.002$, 0.009] on a 59-channel lab cap (12 adults watching a feature film; $9/12$; the
pre-registered skill test failed) and $+0.0005$ [$-0.003$, 0.004] on a 128-channel net (94 children watching movies
from the Healthy Brain Network~\cite{alexander2017hbn}; $48/94$, descriptive).

\section{Statistics of the main-text results}
\label{sup:stats}

\Cref{tab:stats,tab:stats2} list the statistics behind every comparison in \cref{sec:experiments}. Differences are subject means with participant-bootstrap 95\% intervals in brackets; $k/n$ counts subjects
with a positive difference; $p$ is an exact subject-level sign-flip test (15 core subjects) or 100{,}000 random flips
(24 unseen subjects). Pre-registered decisions used skill. The last column names the pre-registered test behind a row
(\cref{sup:prereg}).

\begin{table*}[p]
  \centering
  \caption{\textbf{Statistics of the main-text comparisons, \cref{sec:exp-scaling,sec:exp-video}.} Rows starting with ``vs.'' compare
  \method{} with the named model. Effect: difference in $r$ with its 95\% interval, unless a unit is given or the row
  reads ``in skill''; rel.: in percent of the other model's $r$; $k/n$: subjects ahead. Best linear: the best
  per-subject, pooled or hierarchical ridge model on EEG history, with the video features on trials; on continuous EEG
  it reaches $r=0.272$ at 908, 1{,}142 and 2{,}012\,h. SLv: clip-locked correlation. Test: the pre-registered test that
  decided the row (\cref{tab:prereg,tab:prereg2}); other rows are secondary or descriptive.}
  \label{tab:stats}
  \footnotesize
  \setlength{\tabcolsep}{5pt}
  \begin{tabular}{@{}llllll@{}}
    \toprule
    Comparison & Effect [95\% CI] & rel. & $k/n$ & $p$ & Test \\
    \midrule
    \multicolumn{6}{@{}l}{\emph{Continuous EEG (\cref{sec:exp-scaling}, \cref{fig:scaling})}} \\
    vs.\ best linear, 14\,h & & $-20.7\%$ & 0/15 & & \\
    vs.\ best linear, 56\,h & & $-7.7\%$ & 0/15 & & \\
    vs.\ best linear, 227\,h & & $-2.8\%$ & 2/15 & & \\
    vs.\ best linear, 685\,h & & $+3.5\%$ & 15/15 & $6\times10^{-5}$ & H10 \\
    \quad in skill & $+0.0042$ & & & & \\
    vs.\ best linear, 908\,h & & $+3.4\%$ & 15/15 & & \\
    vs.\ best linear, 1{,}142\,h & & $+4.0\%$ & 15/15 & & \\
    vs.\ best linear, 2{,}012\,h & & $+4.6\%$ & 15/15 & & \\
    Growth of the lead, 908 to 2{,}012\,h & $+0.0032$ [0.0021, 0.0046] & & 15/15 & $10^{-4}$ & H27 \\
    \quad in skill & $+0.0016$ & & & & \\
    \addlinespace
    Linear world model vs.\ \method{} & $-0.024$ & & 0/15 & & H6 \\
    \quad linear world model vs.\ closed-form ridge & lower (\cref{tab:scaling}) & & & & \\
    LSTM and causal Transformer vs.\ best linear & see \cref{tab:scaling} & & 15/15 each & & H36 (Tr.) \\
    MLP vs.\ best linear, 2{,}012\,h & $+0.002$ & & 12/15 & & \\
    iTransformer vs.\ best linear, 2{,}012\,h & $+0.001$ & & 7/15 & n.s. & \\
    vs.\ MLP & $+0.010$ [0.009, 0.012] & & 15/15 & $6\times10^{-5}$ & H54 \\
    vs.\ iTransformer & $+0.012$ [0.010, 0.014] & & 15/15 & $6\times10^{-5}$ & H55 \\
    \midrule
    \multicolumn{6}{@{}l}{\emph{Video trials (\cref{sec:exp-video}, \cref{fig:video}, \cref{tab:video})}} \\
    vs.\ best history + video & $+0.029$ [0.026, 0.031] & $+12.3\%$ & 15/15 & $6\times10^{-5}$ & H21 \\
    \quad in skill & $+0.0103$ & & & & \\
    \quad test clips ahead & & & 285/394 & & \\
    \quad lead at the first second & & $+4\%$ & & & \\
    \quad lead at the tenth second & & $+18\%$ & & & \\
    \quad horizon gained & 3.3\,s [2.2, 4.5] & & 14/15 & & \\
    vs.\ MLP & $+0.018$ [0.016, 0.020] & & 15/15 & & H56 \\
    vs.\ iTransformer & $+0.012$ [0.009, 0.015] & & 15/15 & & H57 \\
    \addlinespace
    150 to 630\,h of trial EEG, \method{} & $+0.012$ & & 13/15 & & H22 \\
    150 to 630\,h of trial EEG, linear & $+0.005$ & & 11/15 & & \\
    SLv, vs.\ best history + video & $-0.058$ & & 2/15 & & H20 (failed) \\
    SLv, 150 to 630\,h & & $+46\%$ & 15/15 & & H19 \\
    \addlinespace
    Swap cost, \method{} & 0.026 & & 15/15 & & \\
    Swap cost, linear & 0.012 & & 15/15 & & \\
    Clip onset only vs.\ pretrained model & $+0.005$ & & 13/15 & & \\
    \quad in skill & nearly all of the gain & & & & \\
    Clip content beyond onset, \method{} & $+0.007$ & & 10/15 & & \\
    \quad in skill & n.s. & & & & \\
    Clip content beyond onset, linear & $+0.008$ & & 14/15 & & \\
    \bottomrule
  \end{tabular}
\end{table*}

\begin{table*}[p]
  \centering
  \caption{\textbf{Statistics of the main-text comparisons, \crefrange{sec:exp-transfer}{sec:exp-ident}}, as in \cref{tab:stats}. Clip: test over
  the 394 test clips (sign-flip for H13, bootstrap for H58). $^\ast$Against a linear model fitted on all of the
  subject's data, a median of 2{,}046 trials (5.7\,h); descriptive.}
  \label{tab:stats2}
  \footnotesize
  \setlength{\tabcolsep}{5pt}
  \begin{tabular}{@{}llllll@{}}
    \toprule
    Comparison & Effect [95\% CI] & rel. & $k/n$ & $p$ & Test \\
    \midrule
    \multicolumn{6}{@{}l}{\emph{Unseen subjects (\cref{sec:exp-transfer}, \cref{fig:transfer})}} \\
    Frozen video model (\cref{fig:transfer}a) vs.\ best linear & $+0.020$ [0.013, 0.027] & $+8.1\%$ & 23/24 & clip $<10^{-5}$ & H13 \\
    \quad in skill & $+0.0083$ & & & & \\
    Retrained model (\cref{fig:transfer}b), all own data & & $+9.7\%$ & 24/24 & & H26 \\
    No-video world model vs.\ best linear & & $+6.8\%$ & 21/24 & & \\
    Video vs.\ no-video world model & $+0.003$ & & & 0.20 & \\
    Average of \method{} and best linear & & $+9.3\%$ & 24/24 & & H14 \\
    \addlinespace
    Swap cost, \method{} & 0.022 & & 22/24 & & \\
    Swap cost, best linear & 0.011 & & 23/24 & & \\
    Swap cost, \method{} over best linear & $2.0\times$ & & 21/24 & & \\
    \addlinespace
    Head from 200 trials (${\approx}33$\,min)$^\ast$ & & $+3.5\%$ & 15/24 & & \\
    Head from 1{,}000 trials$^\ast$ & & & 23/24 & & \\
    Zero-shot vs.\ pooled linear & & $+3.2\%$ & 20/24 & & \\
    \quad in skill & not better & & 8/24 & & H24 (failed) \\
    \midrule
    \multicolumn{6}{@{}l}{\emph{What the model learns (\cref{sec:exp-interp}, \cref{fig:interp}, \cref{tab:encoders})}} \\
    Features with a positive gain & 155/155 & & 15/15 & & \\
    FDR-significant features, continuous / video & 154 / 155 & & & & \\
    Largest band gain, video trials ($\alpha$) & $+0.034$ & $+20\%$ & & & \\
    \quad the same band on continuous EEG & & $+6.8\%$ & & & \\
    Horizon gained, continuous EEG & 1.5\,s [1.1, 1.9] & & & & \\
    \addlinespace
    Gain, farthest fifth of start states ($1.6\times$) & & $+5.2\%$ & & & \\
    Gain, nearest fifth of start states & & $+3.2\%$ & & & \\
    Farthest minus nearest fifth, in skill ($3.8\times$) & $+0.0072$ [0.0057, 0.0088] & & 15/15 & $10^{-4}$ & H37 \\
    \addlinespace
    In-silico vs.\ recorded tuning, $50\times155$ cells & $r=0.57$ & & & perm.\ 0.001 & H38 \\
    \quad features with positive agreement & 47/50 & & & & \\
    \quad strongest feature (CLIP PC1) & $r=0.94$ & & & & \\
    \quad its $\theta$ map; the loudness $\delta$ map & $r=0.96$; 0.95 & & & & \\
    SLv, V-JEPA 2 vs.\ CLIP + DINOv2 & $+35\%$ [25, 47] & & 15/15 & clip $<10^{-4}$ & H58 \\
    \quad SLv of the two & 0.188 vs.\ 0.139 & & & & \\
    SLv, video vs.\ image models & $+0.031$ & & 15/15 & & H59 \\
    \midrule
    \multicolumn{6}{@{}l}{\emph{Identification (\cref{sec:exp-ident}, \cref{fig:retrieval})}} \\
    Trajectory matching, top-10, vs.\ linear & 28.5\% vs.\ 26.5\% & & 11/15 & 0.04 & \\
    \quad \method{} / context mean & $16\times$ / $12.5\times$ chance & & & & \\
    All 10{,}970 viewings as candidates, vs.\ linear & 13.2\% vs.\ 11.9\% & & 10/15 & 0.065 & H34 (failed) \\
    \quad \method{} / context mean & $145\times$ / $90\times$ chance & & & & \\
    Persistence-controlled, vs.\ linear & 5.5\% vs.\ 4.7\% & & 11/15 & 0.15 & H23 (failed) \\
    \quad \method{} vs.\ chance & $2.2\times$ & & 15/15 & & \\
    \bottomrule
  \end{tabular}
\end{table*}

\section{All pre-registered tests}
\label{sup:prereg}
\label{sup:failed}

\Cref{tab:prereg,tab:prereg2} list every pre-registered test run on the data of this paper, H1 to H60, including those the main text does not discuss. Before each round we wrote down its hypotheses,
models, selection procedure, decision rules and the test data it would score, and scored the test data only
afterwards. Unless noted, a test is decided on skill by the rule of \cref{sec:benchmark}; for H12--H14 the subject
criterion was more than half of the 24 (the outcomes are the same under 16/24), and on the lab cap it was 9/12.
Non-inferiority tests require the lower bound of the 95\% interval above the stated margin, and second attempts after
a failed test use $p<0.025$. The identifiers H7, H39 and H40 were never registered. The core subjects' test data were
scored in many rounds, so later results on them are not independent replications; the 24 unseen subjects were first
scored in the confirmatory test (H12--H14) and later in H24--H26, H53 and H53b. Before this corpus existed, pilot
models trained only on the lab-cap data and on the EEG Moments Dataset~\cite{gifford2026emd}, a few hours per
subject, did not beat linear forecasters; H41--H45 repeat the lab-cap comparison with the pretrained model.

Of the 60 tests, 36 were met and 23 failed; H35 was two-sided. Several failures shaped the paper. The first video
model, trained from scratch, lost to the linear model (H4); pretraining on continuous EEG helped (H9) but needed a
second attempt to beat the best linear model (H8, H11). The linear model's video term matched the mean response
across viewers better than the world model's (H18, H20), although the world model's clip-locked response grew with
more data (H19). On unseen subjects' continuous EEG the frozen model beat the best linear model only nominally (H12),
zero-shot transfer lost (H24, H25), and multi-device pretraining transferred slightly worse to new people (H53,
H53b), so the main models are trained on one device. Fine-tuning the video pathway on the 2{,}012-h core with the
908-h recipe was worse (H32, H33), as early stopping ended 7 of 8 runs at epoch 8 before the video pathway
converged. On the lab cap, pretraining turned the pilots' loss into parity but not into a win (H41--H45).
Identification did not separate \method{} from the linear model (H23, H34; \cref{sup:ident}), neither the visual nor
the audio gain was specific to the matching sensory electrodes (H49, H50; \cref{sup:av}), and the calibration
criterion failed on video trials (H60; \cref{sup:calibration}).

\begin{table*}[p]
  \centering
  \caption{\textbf{Pre-registered tests on continuous EEG and on the lab cap.} Differences in skill unless noted,
  favoring the first-named model; $k/n$ counts subjects with a positive difference; linear models as in
  \cref{tab:stats}. $^\ddagger$Pretrained on the same data plus 142\,h from the 128-channel net and 71\,h from the EEG
  Moments Dataset.}
  \label{tab:prereg}
  \footnotesize
  \setlength{\tabcolsep}{4pt}
  \begin{tabular}{@{}lllll@{}}
    \toprule
    Test & Hypothesis & Result & $k/n$ & Outcome \\
    \midrule
    \multicolumn{5}{@{}l}{\emph{Continuous EEG, held-out sessions of the 15 core subjects}} \\
    H1 & Per-subject world model vs.\ per-subject ridge, ${\approx}46$\,h each & $-0.0021$ ($p=0.009$) & 2/15 & failed \\
    H2 & The gap of H1 shrinks from 1 to 48 recordings per subject & $-0.052$ to $-0.002$ & 15/15 & met \\
    \addlinespace
    H3 & World model pooled over the 15 subjects (908\,h) vs.\ per-subject ridge & $+0.0050$ & 15/15 & met \\
    H5 & The pooled world model vs.\ best linear & $+0.0043$ & 14/15 & met \\
    H6 & The pooled world model vs.\ the linear world model & $+0.025$ & 15/15 & met \\
    H6b & H6 with the final 80-epoch schedule for both models & $+0.013$ & 15/15 & met \\
    \addlinespace
    H10 & \method{} vs.\ best linear at 685\,h & $+0.0042$ & 15/15 & met \\
    H29 & \method{} vs.\ best linear at 2{,}012\,h & $+0.0058$ & 15/15 & met \\
    H27 & The lead over best linear grows from 908 to 2{,}012\,h & $+0.0016$ & 15/15 & met \\
    H28 & \method{} at 2{,}012\,h vs.\ at 908\,h & $+0.0017$ & 14/15 & met \\
    H30 & 2{,}012 vs.\ 908\,h of data with the same step budget & $+0.0009$ & 11/15 & met \\
    \addlinespace
    H15 & JEPA-style latent loss vs.\ none at 908\,h & $+0.0010$ & 15/15 & met \\
    H16 & The same at 227\,h & $+0.0011$ & 13/15 & met \\
    H31 & The same at 2{,}012\,h & $+0.0009$ & 15/15 & met \\
    H17 & With the latent loss, \method{} beats best linear already at 227\,h & $-0.0058$ & & failed \\
    \addlinespace
    H35 & Causal Transformer vs.\ GRU core, 2{,}012\,h (two-sided) & $-0.0016$ & 0/15 & GRU better \\
    H36 & Causal Transformer core vs.\ best linear, 2{,}012\,h & $+0.0042$ & 15/15 & met \\
    H54 & \method{} vs.\ the MLP direct forecaster, in $r$ & $+0.010$ & 15/15 & met \\
    H55 & \method{} vs.\ the iTransformer direct forecaster, in $r$ & $+0.012$ & 15/15 & met \\
    \addlinespace
    H37 & Skill gain over best linear, farthest vs.\ nearest fifth of starting states & $+0.0072$ & 15/15 & met \\
    H46 & Multi-device pretraining$^\ddagger$ non-inferior to one device (margin 0.001) & $+0.0001$ [$-0.00002$, 0.0003]
      & & met \\
    H51 & The same in $r$ (margin 0.003) & $+0.0003$ [${\approx}0$, 0.0008] & & met \\
    H60 & Calibration slope in $[0.8, 1.25]$ at all steps in ${\geq}13/15$ subjects (validation, test) & 14/15, 15/15;
      video 6/15 & & failed (video) \\
    \midrule
    \multicolumn{5}{@{}l}{\emph{59-channel lab cap, 12 adults watching a feature film}} \\
    H41 & Fine-tuned \method{} vs.\ best linear & $+0.0022$ ($p=0.13$) & 8/12 & failed \\
    H42 & The same model vs.\ the model trained only on these data & $+0.0136$ & 12/12 & met \\
    H43 & Frozen \method{} + new head vs.\ best linear & $+0.0022$ ($p=0.11$) & 8/12 & failed \\
    H44 & Multi-device$^\ddagger$ vs.\ one-device pretraining, frozen + new head & $+0.0000$ & 5/12 & failed \\
    H45 & Multi-device$^\ddagger$ pretraining vs.\ best linear & $+0.0023$ ($p=0.11$) & 9/12 & failed \\
    \bottomrule
  \end{tabular}
\end{table*}

\begin{table*}[p]
  \centering
  \caption{\textbf{Pre-registered tests on video trials}, of the 15 core subjects and of the 24 unseen subjects, as in
  \cref{tab:prereg}. SLv: clip-locked correlation. $^\dagger$Decided at the clip level (sign-flip over the 394 test
  clips) together with the subject count. $^\ddagger$As in \cref{tab:prereg}. $^\S$One of three seeds stopped at
  epoch 0.}
  \label{tab:prereg2}
  \footnotesize
  \setlength{\tabcolsep}{4pt}
  \begin{tabular}{@{}lllll@{}}
    \toprule
    Test & Hypothesis & Result & $k/n$ & Outcome \\
    \midrule
    \multicolumn{5}{@{}l}{\emph{Video trials, test viewings of the 15 core subjects}} \\
    H4$^\dagger$ & Video model trained from scratch (150\,h) vs.\ per-subject history + video ridge & $-0.004$ & 3/15 &
      failed \\
    H8$^\dagger$ & Video model fine-tuned from the continuous core vs.\ best linear & $+0.0007$ & 8/15 & failed \\
    H11$^\dagger$ & Second attempt at H8 (wider learning-rate grid; clip-level $p<0.025$) & $+0.0047$ (clip $p=0.0019$)
      & 12/15 & met \\
    H9$^\dagger$ & The fine-tuned model vs.\ the model trained from scratch & $+0.0128$ & 15/15 & met \\
    H21$^\dagger$ & \method{} vs.\ best linear, 630\,h & $+0.0103$ & 15/15 & met \\
    H22$^\dagger$ & 630 vs.\ 150\,h of video-paired EEG & $+0.0054$ & 14/15 & met \\
    H32$^\dagger$ & Video model on the 2{,}012-h core vs.\ on the 908-h core & $-0.0042$ & 1/15 & failed \\
    \addlinespace
    H18 & SLv, video model (150\,h) vs.\ best linear & $-0.079$ & 0/15 & failed \\
    H20 & SLv, \method{} (630\,h) vs.\ best linear & $-0.058$ & 2/15 & failed \\
    H19 & SLv, 630 vs.\ 150\,h of video-paired EEG & 0.152 vs.\ 0.104 & 15/15 & met \\
    H33 & SLv, video model on the 2{,}012-h vs.\ the 908-h core & 0.128 vs.\ 0.152 & & failed \\
    H58 & SLv, best visual encoder on validation (V-JEPA 2) vs.\ CLIP + DINOv2 & $+0.049$ [0.035, 0.065] & 15/15 & met \\
    H59 & SLv, video vs.\ image encoders & $+0.031$ [0.024, 0.037] & 15/15 & met \\
    \addlinespace
    H23 & Persistence-controlled identification, top-10 of 394 clips, vs.\ best linear & 5.5\% vs.\ 4.7\% ($p=0.15$) &
      11/15 & failed \\
    H34 & Trajectory matching with all 10{,}970 viewings as candidates, vs.\ best linear & $+1.3$ points ($p=0.065$) &
      10/15 & failed \\
    H38 & In-silico tuning correlates with recorded tuning (validation clips) & $r=0.57$ (perm.\ $p=0.001$) & & met \\
    \addlinespace
    H47 & Visual features improve parieto-occipital $r$ & $+0.0076$ & 11/15 & met \\
    H48 & Audio features improve temporal $r$ & $+0.0014$ & 12/15 & met \\
    H49 & The visual gain is larger at parieto-occipital than at temporal sites & $+0.0023$ ($p=0.074$) & 13/15 & failed \\
    H50 & The audio gain is larger at temporal than at parieto-occipital sites & $-0.0005$ & 7/15 & failed \\
    \addlinespace
    H52 & Multi-device pretraining$^\ddagger$ non-inferior in $r$ (margin 0.003) & $-0.0029$ [$-0.0053$, $-0.0003$] & &
      failed$^\S$ \\
    H52b & Second attempt, early stopping counted from epoch 20 & $+0.0013$ [$-0.0001$, 0.0026] & 11/15 & met \\
    H56 & \method{} vs.\ the MLP direct forecaster, in $r$ & $+0.018$ & 15/15 & met \\
    H57 & \method{} vs.\ the iTransformer direct forecaster, in $r$ & $+0.012$ & 15/15 & met \\
    \midrule
    \multicolumn{5}{@{}l}{\emph{Unseen subjects, test viewings of the 24 subjects (H12: continuous EEG)}} \\
    H12 & Frozen continuous model + new head vs.\ best linear & $+0.0015$ ($p=0.067$) & 15/24 & failed \\
    H13$^\dagger$ & Frozen video model + new head vs.\ best linear (\cref{fig:transfer}a) & $+0.0083$ & 23/24 & met \\
    H14$^\dagger$ & The average of the two forecasts beats both & $+0.0106$ / $+0.0023$ & 24/24, 18/24 & met \\
    H26 & Model retrained on the other 41 subjects + new head vs.\ best linear & $+0.0097$ & 24/24 & met \\
    H24 & The same model, zero-shot, vs.\ pooled linear & $-0.0043$ & 8/24 & failed \\
    H25 & The same, persistence-controlled identification & $-0.58$ points ($p=0.11$) & 6/24 & failed \\
    H53 & Multi-device pretraining$^\ddagger$ non-inferior in $r$ (margin 0.003) & $-0.0018$ [$-0.0058$, 0.0022] & 10/24 &
      failed \\
    H53b & Second attempt, three seeds & $-0.0020$ [$-0.0053$, 0.0012] & 11/24 & failed \\
    \bottomrule
  \end{tabular}
\end{table*}

\clearpage